\documentclass{article}

\usepackage[preprint]{neurips_2023}

\usepackage[utf8]{inputenc} 
\usepackage[T1]{fontenc}    
\usepackage{hyperref}       
\usepackage{url}            
\usepackage{booktabs}       
\usepackage{amsfonts}       
\usepackage{nicefrac}       
\usepackage{microtype}      
\usepackage{xcolor}         
\usepackage{amsmath}
\usepackage{multicol}
\usepackage{multirow}
\usepackage{color, colortbl}
\definecolor{Gray}{gray}{0.9}
\newcolumntype{g}{>{\columncolor{Gray}}c}
\usepackage{pifont}
\newcommand{\cmark}{\text{\ding{51}}}
\newcommand{\xmark}{\text{\ding{55}}}
\usepackage{graphicx}
\graphicspath{ {./images/} }

\usepackage{natbib}
\setcitestyle{square,numbers,sort}

\title{{DARTS}: Double Attention Reference-based Transformer for Super-resolution}

\author{%
  Masoomeh Aslahishahri \\
  Department of Computer Science\\
  University of Saskatchewan\\
  Saskatoon, Canada \\
  \texttt{masi.aslahi@usask.ca} \\
  \And
  Jordan Ubbens \\
  Department of Computer Science \\
  University of Saskatchewan\\
  Saskatoon, Canada \\
  \texttt{jordan.ubbens@usask.ca} \\
  \AND
  Ian Stavness \\
  Department of Computer Science \\
  University of Saskatchewan\\
  Saskatoon, Canada \\
  \texttt{ian.stavness@usask.ca} \\
}

\begin{document}

\maketitle

\begin{abstract}
  We present DARTS, a transformer model for reference-based image super-resolution. DARTS learns joint representations of two image distributions to enhance the content of low-resolution input images through matching correspondences learned from high-resolution reference images. Current state-of-the-art techniques in reference-based image super-resolution are based on a multi-network, multi-stage architecture. In this work, we adapt the double attention block from the GAN literature, processing the two visual streams separately and combining self-attention and cross-attention blocks through a gating attention strategy. Our work demonstrates how the attention mechanism can be adapted for the particular requirements of reference-based image super-resolution, significantly simplifying the architecture and training pipeline. We show that our transformer-based model performs competitively with state-of-the-art models, while maintaining a simpler overall architecture and training process. In particular, we obtain state-of-the-art on the SUN80 dataset, with a PSNR/SSIM of 29.83 / .809. These results show that attention alone is sufficient for the RSR task, without multiple purpose-built subnetworks, knowledge distillation, or multi-stage training.
\end{abstract}

\section{Introduction}
Image super resolution (SR) refers to enhancing the spatial resolution of a low-resolution (LR) image, transforming it into a high-resolution (HR) image, while recovering crucial and realistic textures and details \cite{zheng2018crossnet, zhang2019image}. Image super resolution can improve the user experience of digital media content. For example, SR can enhance the texture of video frames in computer games without increasing the computational cost, or improve the digital zoom functionality of smartphones. A wide range of computer vision tasks such as medical imaging and remote sensing can also significantly benefit from image super resolution \cite{li2021review}. The super resolution problem is generally divided into two sub-problems: single image super resolution (SISR) and reference-based image super resolution (RSR). Using classical approaches, SISR produces blurry images with aliasing artifacts because significant degradations occur when images are downsampled. However, recent SISR models have tried to enhance the image SR process and produce images as close as possible to the corresponding target HR images \cite{dai2019second, kim2016accurate, kim2016deeply, lim2017enhanced, conde2023swin2sr, liang2021swinir}. 

\begin{figure}[t]
\centering
\includegraphics[width=.95\textwidth]{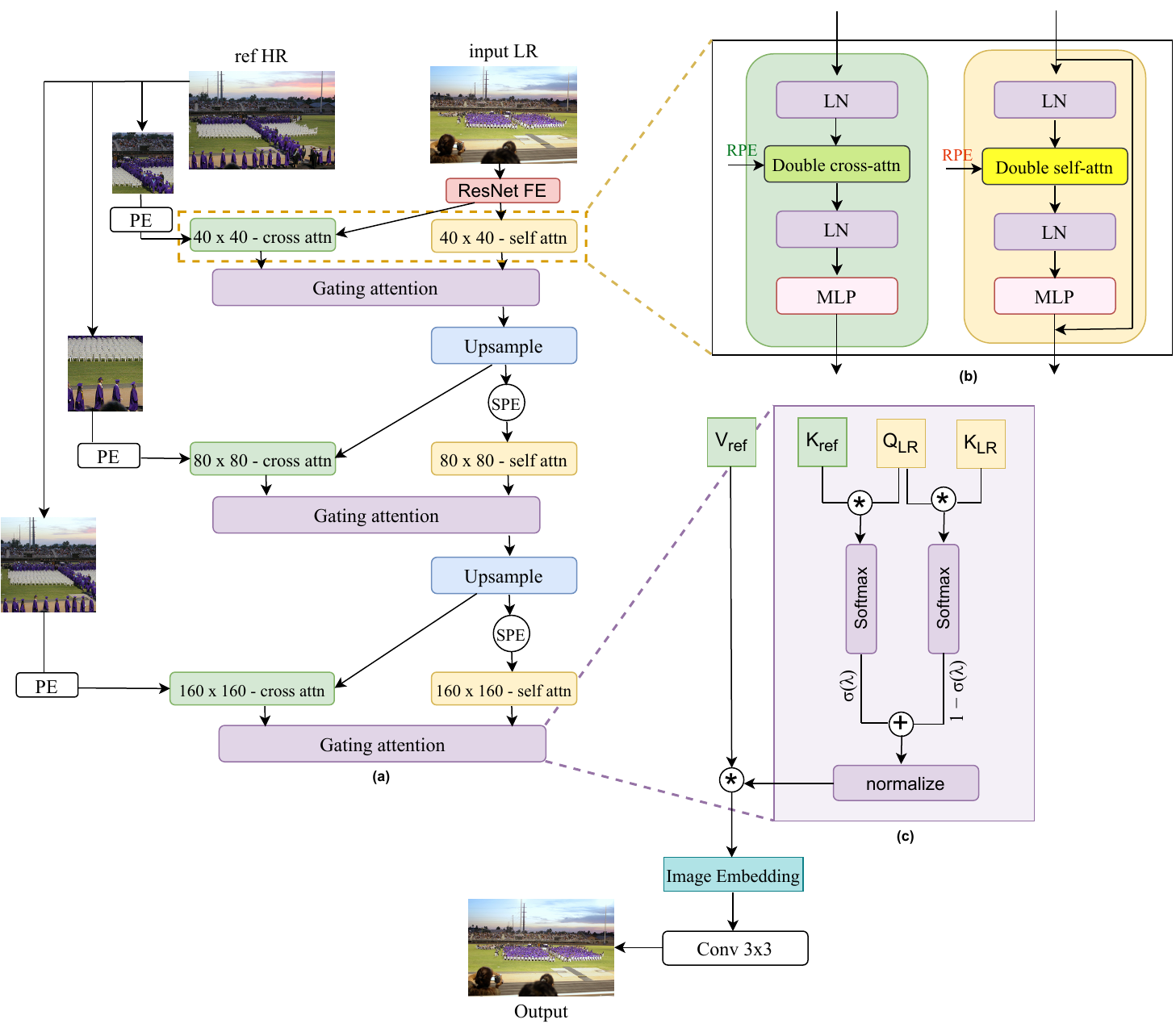}
\caption{Overview of DARTS. a) The proposed architecture takes two images as input, where LR features are derived from an adapted ResNet feature extractor (FE) and HR reference features are derived from a patch embedding (PE) module. b) Double attention is used across the two image distributions, using both self- and cross-attention blocks. c) Gating attention balances self- and cross-attention matrices, modulated by a gating parameter $(\lambda)$.}
\label{overview}
\end{figure}

The latest progress in image SR has been made by RSR, where HR reference images are used to transfer HR textures to corresponding LR images to produce more realistic images preserving fine-grained textures \cite{yang2020learning}. This texture transfer is a challenging problem because it requires finding matching correspondences between the LR input image and the HR reference image. There are two major challenges inherent in this task: 1) the spatial (transformation) gap between similar regions in the LR input and HR reference images, where corresponding landmarks appear at different positions and orientations in the two images, and 2) the resolution gap between the LR input and HR reference images. For the transformation gap, the input image and the reference image may be very similar in their content, but the appearance of objects may differ due to geometrical transformations in scale, position and shape of objects, which can lead to suboptimal texture transfer. For the resolution gap, the amount of information between the LR input and HR reference images is significantly imbalanced because some information is missing in the LR input image, which can hinder the correspondence matching process, especially in fine-grained texture regions.

Recent successes in RSR have adopted self-supervised learning \cite{cao2022reference} or a coarse-to-fine matching module \cite{lu2021masa} to produce quality HR images. Current state-of-the-art (SOTA) techniques are built on the general architecture which was first presented in $C^2$-matching \cite{jiang2021robust}. These techniques involve a contrastive correspondence module, trained via student-teacher distillation, which extracts correspondences from the input and reference images. These correspondences are then transformed by a dynamic aggregation module followed by a restoration module which outputs the super resolved image. While the correspondences were originally extracted by a convolutional backbone, subsequent papers have explored other options such as deformable transformers \cite{cao2022reference}. Although the correspondence module is interchangeable, these methods tend to maintain the same general structure as $C^2$-matching with multiple subnetworks and multiple stages of training.

In this work, we present a model called DARTS, extending double attention to jointly learn the matching correspondences of two image distributions using self-attention and cross-attention blocks. Our key technical innovation is to introduce separate streams across two image distributions that communicate through an adaptive gating attention strategy \cite{d2021convit} attending to the content of self-attention and cross-attention blocks, regulating the attention paid to each block by adjusting a gating parameter. This structure can accommodate finding the matching correspondences between the two image distributions and provide interaction between them at varying representation sizes. In our ablation studies, we demonstrate that this structure outperforms double attention used in a single-stream model. In contrast to current SOTA methods such as $C^2$-matching \cite{jiang2021robust} and DATSR \cite{cao2022reference}, the entire network consists of a single module which is trained end-to-end using gradient descent.

To the best of our knowledge, our proposed framework is a novel technique in RSR that can achieve quantitative results which are comparable to other SOTA techniques in the literature, while using a simplified architecture and training strategy. The main contributions of this study are:
\begin{itemize}
    \item Using a modified double attention module to learn joint representations across two image distributions and predict correspondences. This allows the attention mechanism to transfer fine-grained textures from the HR reference image to the corresponding LR image, while being robust to differences in the shape and scale of objects. 
    \item We use a gating attention strategy to simultaneously attend to the content of self-attention and cross-attention blocks. This technique provides the attention heads with freedom to modulate the combination of features in each transformer block by adjusting a gating parameter.
    \item We empirically show that our framework achieves comparable or better results than previous methods with more complex training strategies. On the Sun80 dataset, our approach achieves a PSNR of $29.83 ~dB$, outperforming all current RSR methods including those based on $C^2$-Matching \cite{jiang2021robust}.
\end{itemize}

\textbf{Reproducibility.} An open-source implementation of our technique, pre-trained models, and output images are available at \url{https://github.com/bia006/DARTS}.

\section{Related Work}

\textbf{Single Image Super Resolution.}
SISR is a common problem in computer vision and has been studied for an extended period of time. SISR aims to enhance the spatial details within the downsampled LR image and super-resolve the LR image to the corresponding HR target image. The first deep learning-based SISR study used an interpolated LR image with three-layer CNN to represent the mapping function between the LR and target HR images \cite{dong2015image}. They later improved the SISR process by using a deconvolutional layer to upsample the feature map to the desired size \cite{dong2016accelerating}. Later, deeper networks employing residual learning and dense skip connections were proposed for image SR \cite{kim2016accurate, kim2016deeply, tong2017image, lim2017enhanced, liu2020residual}. The checkerboard artifacts caused by the deconvolutional layer were reduced by replacing it with subpixel convolutional layers to upsample the feature map size \cite{shi2016real}. To improve the performance of SISR models, a channel attention block was proposed to explore inter-channel correlations \cite{zhang2018image}. Different research adapted non-local attention to model long-range dependencies to reproduce quality SR images \cite{dai2019second, liu2018non, zhang2019residual, mei2020image, mei2021image}. Perceptual loss and MSE were introduced to overcome the overly smoothed textures in PSNR-oriented methods \cite{johnson2016perceptual, simonyan2014very}. Generative adversarial networks (GAN) \cite{goodfellow2020generative} were used in image SR models \cite{ledig2017photo} and witnessed further refinement in other studies \cite{sajjadi2017enhancenet, wang2018esrgan, zhang2019ranksrgan}. With the advent of vision transformers, deeper neural networks attending to attention mechanism have been recently introduced to produce higher quality images while preserving textures \cite{conde2023swin2sr, liang2021swinir}. 

\textbf{Reference-based Image Super Resolution.}
The most significant difference between the SISR and RSR models is that the RSR model receives additional information for image SR in the form of a HR reference image. The additional HR reference image along with the LR input image can improve the quality of the reconstructed images. The texture representations significantly missing in the LR input image can be transferred from the corresponding HR reference image which contains the same or similar content. In \cite{zhang2019image}, a multi-scale feature transformation was proposed to fuse swapped features obtained from local patch matching into the LR input image. A coarse-to-fine matching scheme was introduced to reduce the computational complexity while boosting the spatial feature matching when there is a domain shift problem between the LR input and HR reference images \cite{lu2021masa}. Elsewhere in the literature, a contrastive learning network was used to match relevant correspondences between the LR input and HR reference images \cite{jiang2021robust}. 

$C^2$-matching is a SOTA model for RSR \cite{jiang2021robust}. The $C^2$-matching process consists of a two main training stages -- one to learn correspondence matching and a second to intake correspondences and synthesize a super-resolved output image. The first stage consists of a contrastive learning module with teacher-student distillation, while the second stage uses a dynamic aggregation module and a restoration network. In the original work, all of the subnetworks are based on CNNs. Other SOTA models have been introduced, borrowing significantly from the structure of $C^2$-matching while making changes to improve its feature matching capabilities. In \cite{cao2022reference}, a deformable attention block built on UNet \cite{ronneberger2015u} was introduced to attend to the content of feature encoders. A refinement module was also proposed to select/align features for better performance \cite{zhang2022rrsr}. 

\textbf{StyleSwin.} StyleSwin \cite{zhang2022styleswin} is a Swin transformer-based \cite{liu2021swin} GAN model for generating synthetic images. To enlarge the receptive field of Swin transformer blocks, StyleSwin introduced the \textit{double attention} mechanism, applying attention across both local and shifted windows. The authors find that this significantly improves the quality of generated images.

\section{Approach}

\subsection{Problem Setup}
In this work, we adapt double attention from image generation to RSR. While the original application involved computing self-attention in the local and shifted windows for a single input in order to broaden the network's receptive field, we adapt it to learn matching correspondences from the HR reference image to the LR input counterpart.

We denote the input LR, the corresponding HR reference, the reconstructed HR, and the HR target images as $I_{LR}$, $I_{ref}$, $I_{SR}$ and $I_{HR}$, respectively. Each image distribution contains $N$ samples. 
\subsection{Model Architecture}
The architecture of DARTS is illustrated in Figure \ref{overview}. The network receives the LR image and HR reference image as inputs and upsamples the LR input image by a factor of four through a cascade of Swin transformer blocks to enhance the spatial resolution.

To account for interaction across adjacent windows, the Swin transformer uses shifted window partitioning in every other block. We adapt this strategy across two image distributions for every Swin block. Given the input feature maps $I_{LR}^l \in \mathbb{R}^{H \times W \times C} $ and $I_{ref}^l \in \mathbb{R}^{H \times W \times C}$ of layer $l$, the following Swin blocks operate as:
\begin{equation}
    \text{\footnotesize{Local windows}} = \small \begin{cases}
    \hat{I_{LR}}^{l} = \textrm{W-MHA}(\textrm{LN}(I_{LR}^l)) + I_{LR}^l, \quad \quad \hat{I_{ref}}^{l} = \textrm{W-MHA}(\textrm{LN}(I_{ref}^l)) + I_{ref}^l\\
    I_{LR}^{l+1} = \textrm{MLP}(\textrm{LN}(\hat{^I_{LR}}^l)) + \hat{I_{LR}}^l, \quad \quad \quad I_{ref}^{l+1} = \textrm{MLP}(\textrm{LN}(\hat{I_{ref}}^l)) + \hat{I_{ref}}^l 
   \end{cases} 
\end{equation}
and, 
\begin{equation}
    \text{\footnotesize{Shifted windows}} = \small \begin{cases}
     \hat{I_{LR}}^{l+1} = \textrm{SW-MHA}(\textrm{LN}(I_{LR}^{l+1})) + I_{LR}^{l+1}, \quad \quad \hat{I_{ref}}^{l+1} = \textrm{SW-MHA}(\textrm{LN}(I_{ref}^{l+1})) + I_{ref}^{l+1}\\
    I_{LR}^{l+2} = \textrm{MLP}(\textrm{LN}(\hat{I_{LR}}^{l+1})) + \hat{I_{LR}}^{l+1}, \quad \quad \quad
    I_{ref}^{l+2} = \textrm{MLP}(\textrm{LN}(\hat{I_{ref}}^{l+1})) + \hat{I_{ref}}^{l+1}
    \end{cases}
\end{equation}
where W-MHA and SW-MHA represent window-based multi-head attention across the local and shifted window partitioning respectively, and LN denotes layer normalization. Since the computational weight of the Swin block corresponds linearly to image size, the network is scalable to higher resolution images. For the RSR task, we extend the Swin transformer to learn joint representations across two image distributions, processing $I_{LR}$ and $I_{ref}$ inputs in separate streams and computing self-attention and cross-attention matrices. The two streams then interact through a gating attention strategy.

A 16-block ResNet \cite{he2016deep} with SpectralNorm \cite{miyato2018spectral} normalization is used to extract image features from a $40 \times 40$ LR input image and this feature map is used as the input to the first transformer block. For the corresponding reference image, unlike the other SOTA methods which use dense features extracted using a pre-trained VGG network \cite{cao2022reference, jiang2021robust,zhang2019image}, we use random crops from the original HR reference image without spatial loss. The random crops are $(40 \times 40), (80 \times 80)$ and $(160 \times 160)$ spatially, matching the input size of each transformer block. We adapt the image patching module from ViT \cite{dosovitskiy2020image} to extract the feature representations out of the cropped patches. Experiments incorporating VGG features extracted from the $relu3-1, relu2-1$ and $relu1-1$ layers as an input in each transformer block instead of random patches yielded similar results (data not shown).  

\textbf{Double Attention.}
We use local and shifted window partitioning across the two inputs $(I_{LR}, I_{ref})$ in each transformer block, which enlarges the receptive field by $2.5k$ across each input image, where $k \times k$ denotes the window size (i.e. $k = 8$). This strategy allows the network to attend to the context of four windows (two local and two shifted windows) at the same time, containing $I_{LR}$ and $I_{ref}$ representations. We compute the query and key from $I_{LR}$ for self-attention and the query from $I_{LR}$ and key from $I_{ref}$ for cross-attention blocks. The network alternates between the local and shifted windows across the two distributions, first attending to the local window on $I_{LR}$ and computing attention scores for the self-attention block, and then attending to the context of the local window on $I_{LR}$ and the context of shifted window on $I_{ref}$ to perform cross-attention. Next, the network alternates between local and shifted windows across the two distributions to compute the scores from the self- and cross-attention blocks. In every step, we compute two attention matrices containing self- and cross-attention scores which are potentially of different magnitudes. To avoid this difference in magnitudes resulting in one attention matrix being overpowered by the other, we use a gating attention block to balance their contributions. We denote $I_w$ and $I_{sw}$ to represent non-overlapping patches under the local and shifted window partitioning respectively as 
\begin{equation}
    \small
    head_i =
    \begin{cases}
    \text{Self-Attn}(I_{LR_w}W_i^Q, I_{LR_w} W_i^K, I_{ref_w} W_i^V), \\\text{Cross-Attn}(I_{LR_{sw}} W_i^Q, I_{ref_{sw}} W_i^K, I_{ref_{sw}} W_i^V)\\
    \text{Self-Attn}(I_{LR_{sw}}W_i^Q, I_{LR_{sw}} W_i^K, I_{ref_{sw}} W_i^V), \\\text{Cross-Attn}(I_{LR_{w}} W_i^Q, I_{ref_{w}} W_i^K, I_{ref_{w}} W_i^V)\\ 
    \end{cases}
\end{equation}
where $W_i^Q, W_i^K, W_i^V \in \mathcal{R}^{C \times C}$ are the query, key and value projection matrices for $i^{th}$ head respectively. 

This technique allows the network to capture larger context more efficiently, i.e. with a window size of $k = 8 \times 8$ and an input feature map of $80 \times 80$, $5$ transformer blocks are required to span the entire feature map, instead of $10$ transformer blocks.

The main architecture uses Swin transformer blocks ($i$), receiving intermediate visual representations $I_{LR}^{i}$ and $I_{ref}^{i}$, as illustrated in Figure \ref{overview}(b). The module computes query, key and value matrices as block inputs and we perform self-attention and cross-attention attending to the content of $I_{LR}^i$ and $I_{ref}^i$ in each block.

For self-attention, the query $q_i$ and key $k_j$ extracted from the $I_{LR}^i$ distribution are passed as inputs to a multi-head attention block, where the attention block produces weighted feature vectors for $I_{LR}^i$ conditioned on itself -- in effect attending to different parts of the same input data in a visual stream in each transformer block. For cross-attention, the query $q_i$ extracted from the $I_{LR}^i$ distribution and the key $k_j$ extracted from the $I_{ref}^i$ distribution are passed as inputs to a multi-head attention block, where the attention block produces weighted feature vectors for $I_{LR}^{i}$ conditioned on $I_{ref}^{i}$ -- in effect performing $I_{ref}^{i}$-conditioned $I_{LR}^{i}$ attention attending to different parts of $I_{ref}^i$ in another visual stream in each transformer block. This strategy simulates the attention mechanism introduced in vision-and-language models \cite{lu2019vilbert}. 

The self-attention and cross-attention matrices are combined through a gating attention mechanism. The value $v_i$ vector is extracted from the $I_{ref}^i$ distribution and the rest of the attention block proceeds with residual learning with the initial $I_{LR}^{i}$ representations.

\textbf{Gating Attention.} Building on the insights of \cite{d2021convit}, we use a gating attention strategy to combine the content of self-attention and cross-attention blocks simultaneously, as illustrated in Figure \ref{overview}(c). We initialize a gating block where each attention head $h$ maintains a learnable $\lambda_h$ \textit{gating parameter} regulating the attention paid to the self-attention scores versus the cross-attention scores. The \textit{gating attention} layer sums the content of self-attention and cross-attention \textit{after} softmax using the \textit{gating parameter} $\lambda_h$ for each attention head. A sigmoid function is used to keep the gating parameter in distribution space. The attention matrix $A^h$ is given by 

\begin{equation}
\begin{split}
    \text{self-attn} &\;=\; \left[ \text{softmax} \left(\frac{Q_{LR}K_{LR}^T}{\sqrt{D_h}}\right) \right] \quad , \\
    \text{cross-attn} &\;=\;  \left[ \text{softmax} \left(\frac{Q_{LR}K_{ref}^T}{\sqrt{D_h}}\right) \right] \quad , \\
    A^h \;=\;  &(1 - \sigma(\lambda_h)) \; \text{self-attn} + \sigma(\lambda_h) \; \text{cross-attn} 
\end{split}
\end{equation}

\noindent where $ \left[ \; \cdot \; \right]$ is a normalization operation and $\sigma$ denotes the sigmoid function. We initialize the gating parameter $\lambda_h$ to 1 as suggested in \cite{d2021convit}.

The gating strategy adjusts the importance of each attention score, in which each attention head projects the most important representation out of the two attention blocks. The final output is 
\begin{equation}
    \text{attn} = \text{concat}(head_0, ..., head_h)W^O
\end{equation}
\noindent where $W^O \in \mathcal{R}^{C \times C}$ denotes the projection matrix from concatenated heads to output. 

\textbf{Positional Encoding.} We use both local and global positional encodings. For the local positional encoding, we use relative positional encoding (RPE) which encodes the spatial information via learnable parameters interacting with queries and keys in each attention block \cite{liu2021swin}. For the global positional encoding we use sinusoidal position encoding (SPE), encoding positional information as a mix of sine and cosine functions \cite{choi2021toward, vaswani2017attention, xu2021positional}, on each upsampling block to provide translation invariance. In practice, RPE is applied within each transformer block, and SPE is applied on each upsampling block informing the global position. 

\textbf{Architectural Details.}
Our framework slices input images of size $M \times M$ into $M/k \times M/k$ non-overlapping patches of $k \times k$ pixels and embeds them into vectors of dimension $D_{emb} = 96N_h$, where $M \times M$, $k \times k$ and $N_h$ denote the input feature map size, window size, and number of attention heads, respectively. The patches are propagated through 3 blocks while keeping dimensionality constant. Each block contains a double attention mechanism across $I_{LR}$ and $I_{ref}$, performing self-attention and cross-attention. Each attention operation is followed by a 2-layer MLP with GeLU activation and a residual connection to the output. Gating attention is applied at the end of each transformer block.
\subsection{Objective Functions}
The objective functions of the proposed method include:

\textbf{Reconstruction Loss.} We use $l_1$ reconstruction loss with $\text{weight} = 10$.:
\begin{equation}
    L_{rec} = ||I_{HR} - I_{SR}||_1
\end{equation}
\textbf{Perceptual Loss.} We employ perceptual loss \cite{johnson2016perceptual} to enhance the visual quality of the reconstructed images, given by
\begin{equation}
    L_{per} = \frac{1}{V}\sum_{i=1}^C||\Phi_i(I_{HR}) - \Phi_i(I_{SR})||_F 
\end{equation}
\noindent where $C$ and $V$ denote the channel number and volume of the feature maps. $\Phi$ represents the $relu5-1$ features out of the VGG19 network \cite{simonyan2014very}. We use $1e-4$ as the weight for $L_{per}$.

\textbf{Adversarial Loss.} We perform adversarial training with the discriminator architecture used in StyleSwin \cite{zhang2022styleswin}. We use hinge \cite{lim2017geometric} with $R_1$ gradient penalty as the adversarial loss function \cite{karras2020analyzing}. We employ balanced consistency regularization (bCR) \cite{zhao2021improved} in training with equal weights on the real and fake images. The weight for the adversarial loss is $1e-4$. We also directly adopt the Wavelet discriminator from StyleSwin \cite{zhang2022styleswin} to combat blocking artifacts. 
\begin{table*}[t!]
\small
\centering
\begin{tabular}{c| l g g g g g} 
 \hline
 \rowcolor{white}
 & Method & CUFED5 & SUN80 & Urban100 & Manga109 & WR-SR \\ 
 \hline\hline
 \rowcolor{white}
 & SRCNN \cite{dong2015image} & 25.33 / .745 & 28.26 / .781 & 24.41 / .738 & 27.2 / .850 & 27.27 / .767\\ 
 \rowcolor{white}
 & EDSR \cite{lim2017enhanced} & 25.93 / .777 & 28.52 / .792 & 25.51 / .783 & 28.93 / .891 & 28.07 / .793 \\
\rowcolor{white}
 & RCAN \cite{zhang2018image} & 26.06 / .769 & 29.86 / .810 & 25.42 / .768 & 29.38 / .895 & 28.25 / .799\\
 \rowcolor{white}
SISR &  SwinIR \cite{liang2021swinir} & 26.62 / .790 & 30.11 / .817 & 26.26 / .797 & 30.05 / .910 & 28.06 / .797\\
& ESRGAN \cite{wang2018esrgan} & 21.90 / .633 & 24.18 / .651 & 20.91 / .620 & 23.53 / .797 & 26.07 / .726 \\ 
\rowcolor{white}
& ENet \cite{sajjadi2017enhancenet} & 24.24 / .695 & 26.24 / .702 & 23.63 / .711 &  25.25 / .802 & 25.47 / .699 \\
& RankSRGAN \cite{zhang2019ranksrgan} & 22.31 / .635 & 25.60 / .667 & 21.47 / .624 & 25.04 / .803 & 26.15 / .719 \\
\rowcolor{white}
\hline\hline
\rowcolor{white}
& CrossNet \cite{zheng2018crossnet} & 25.48 / .764 & 28.52 / .793 & 25.11 / .764 & 23.36 / .741 & -\\
& SRNTT \cite{zhang2019image} & 25.61 / .764 & 27.59 / .756 & 25.09 / .774 & 27.54 / .862 & 26.53 / .745\\
\rowcolor{white}
& SRNTT-\textit{rec} \cite{zhang2019image} & 26.24 / .784 & 28.54 / .793 & 25.50 / .783 & 28.95 / .885 & 27.59 / .780\\
& MASA \cite{lu2021masa} & 24.92 / .729 & 27.12 / .708 & 23.78 / .712 & 27.44 / .849 & 25.76 / .717\\
\rowcolor{white}
& MASA-\textit{rec} \cite{lu2021masa} & 27.54 / .814 & 30.15 / .815 & 26.09 / .786 & 30.28 / .909 & 28.19 / .796\\
& TTSR \cite{yang2020learning} & 25.53 / .765 & 28.59 / .774 & 24.62 / .747 & 28.70 / .886 & 26.83 / .762\\
\rowcolor{white}
& TTSR-\textit{rec} \cite{yang2020learning} & 27.09 / .804 & 30.02 / .814 & 25.87 / .784 & 30.09 / .907 & 27.97 / .792\\
RSR & $C^2$-Matching \cite{jiang2021robust} & 27.16 / .805 & 29.75 / .799 & 25.52 / .764 & 29.73 / .893 & 27.80 / .780\\
\rowcolor{white}
& $C^2$-Matching-\textit{rec} \cite{jiang2021robust} & 28.24 / .841 & 30.18 / .817 & 26.03 / .785 & 30.47 / .911 & 28.32 / .801\\
& DATSR \cite{cao2022reference} & 27.95 / .835 & 29.77 / .800 & \textcolor{red}{25.92} / \textcolor{red}{.775} & 29.75 / .893 & 27.87 / \textcolor{red}{.787}\\
\rowcolor{white}
& DATSR-\textit{rec} \cite{cao2022reference} & 28.72 / .856 & 30.20 / .818 & \textcolor{blue}{26.52} / \textcolor{blue}{.798} & 30.49 / .912 & 28.34 / \textcolor{blue}{.805}\\
& RRSR \cite{zhang2022rrsr} & \textcolor{red}{28.09} / \textcolor{red}{.835} & 29.57 / .793 & 25.68 / .767 & \textcolor{red}{29.82} / .893 & \textcolor{red}{27.89} / .784\\
\rowcolor{white}
& RRSR-\textit{rec} \cite{zhang2022rrsr} & \textcolor{blue}{28.83} / \textcolor{blue}{.856} & \textcolor{blue}{30.13} / \textcolor{blue}{.816} & 26.21 / .790 & \textcolor{blue}{30.91} / \textcolor{blue}{.913} & \textcolor{blue}{28.41} / .804\\
& DARTS (ours) & 26.6 / .781 & \textcolor{red}{29.83} / \textcolor{red}{.809}  & 25.6 / .772 & 29.8 / \textcolor{red}{.898} & 27.78 / \textcolor{red}{.787}\\
\rowcolor{white}
& DARTS-\textit{rec} (ours) & 26.4 / .781 & 29.9 / .81  & 25.51 / .770 & 29.8 / .901 & 27.8 / .786\\
\hline
\end{tabular}
\caption{Quantitative comparisons using PSNR / SSIM metrics. The SISR and RSR methods are grouped accordingly. The \textit{`-rec'} denotes only reconstruction ($l_1$) loss. The methods shaded in grey use a GAN loss. The highest values for networks trained with multiple loss functions are shown in \textcolor{red}{red} and the highest values for networks trained with only reconstruction ($l_1$) loss are shown in \textcolor{blue}{blue}.}
\label{quantitative_comp}
\end{table*}
\section{Experiments}
\subsection{Experimental Settings}
\textbf{Datasets}. Performance was evaluated using CUFED5 \cite{zhang2019image}, SUN80 \cite{sun2012super}, Urban100 \cite{huang2015single}, and Manga109 \cite{matsui2017sketch}. For CUFED5, the training set contains $11,871$ pairs of input and reference images. The CUFED5 test set contains $126$ images, in which every input image is paired with five reference images from different levels of similarity. The input image and its respective five references have been selected from an album of the same event. The Webly-Referenced SR dataset \cite{jiang2021robust} consists of $80$ image pairs where each pair contains an input image and reference image. The SUN80 dataset contains $80$ images in which each input image has $20$ reference images. For the Urban100 and Manga109 datasets, which are SISR datasts, we adopt the same evaluation strategy as \cite{yang2020learning, zhang2019image}. These datasets consist of $100$ and $109$ images respectively and a random image from the same dataset is used as a reference image. The LR input images are created by downsampling the HR target images $4\times$ using bicubic interpolation.

\textbf{Evaluation Metrics}. Peak-Signal-to-Noise Ratio (PSNR) and Structural Similarity Index (SSIM) \cite{wang2004image} are used as evaluation metrics. The Y channel of YCrCb colorspace is used for measuring the PSNR and SSIM values on each dataset. 

\subsection{Implementation Details}
\label{sec:implementation}
For training, we use the Adam solver \cite{kingma2014adam} with $\beta_1=0.0, \beta_2=0.99$. We use a single-cycle learning rate schedule  \cite{smith2019super}, with a maximum learning rate of $1e-4$. During training, we use a batch size of 4 with one sample per GPU. At inference time we use $(192 \times 192)$, $(384 \times 384)$, and $(768 \times 768)$ patches from the reference image.
\subsection{Results and Analysis}
 We compare the proposed technique with recent CNN- and transformer-based networks from the literature. For SISR methods, we include SRCNN \cite{dong2015image}, EDSR \cite{lim2017enhanced}, RCAN \cite{zhang2018image}, SwinIR \cite{liang2021swinir}, ESRGAN \cite{wang2018esrgan}, ENet \cite{sajjadi2017enhancenet} and RankSRGAN \cite{zhang2019ranksrgan}. For RSR methods, we include CrossNet \cite{zheng2018crossnet}, SRNTT \cite{zhang2019image}, MASA \cite{lu2021masa}, TTSR \cite{yang2020learning}, $C^2$-Matching \cite{jiang2021robust}, DATSR \cite{cao2022reference} and RRSR \cite{zhang2022rrsr}.
 
\textbf{Quantitative results.} Table \ref{quantitative_comp} shows a quantitative comparison of DARTS against existing SOTA methods. The proposed method achieves results which are competitive with SOTA on many datasets, and achieves SOTA in the SUN80 benchmark. Methods using the general $C^2$-matching architecture \cite{cao2022reference, zhang2022rrsr} still achieve the best performance in the majority of datasets -- however, these models require a more complex architecture and training procedure to achieve these results.  

\begin{figure}[tb!]
\centering
\scriptsize
\begin{tabular}{ccccc ccccc}
\multicolumn{2}{c}{input image} & LR & ESRGAN & RankSRGAN & \multicolumn{2}{c}{input image} & LR & ESRGAN & RankSRGAN\\ \hline
\multicolumn{2}{c}{reference image} & SRNTT & DATSR & DARTS (ours) & \multicolumn{2}{c}{reference image} & SRNTT & DATSR & DARTS (ours)\\
\multicolumn{2}{c}{\includegraphics[width=.15\linewidth,height=1cm]{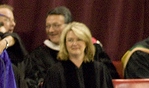}}&
\multicolumn{1}{c}{\includegraphics[width=.075\linewidth,height=1cm]{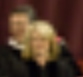}}&
\multicolumn{1}{c}{\includegraphics[width=.075\linewidth,height=1cm]{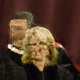}}&
\multicolumn{1}{c}{\includegraphics[width=.075\linewidth,height=1cm]{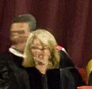}}&
\multicolumn{2}{c}{\includegraphics[width=.15\linewidth,height=1cm]{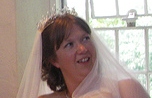}}&
\multicolumn{1}{c}{\includegraphics[width=.075\linewidth,height=1cm]{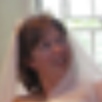}}&
\multicolumn{1}{c}{\includegraphics[width=.075\linewidth,height=1cm]{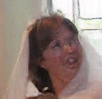}}&
\multicolumn{1}{c}{\includegraphics[width=.075\linewidth,height=1cm]{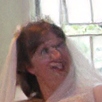}}
\\
\multicolumn{2}{c}{\includegraphics[width=.15\linewidth,height=1cm]{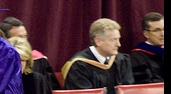}}&
\multicolumn{1}{c}{\includegraphics[width=.075\linewidth,height=1cm]{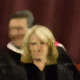}}&
\multicolumn{1}{c}{\includegraphics[width=.075\linewidth,height=1cm]{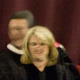}}&
\multicolumn{1}{c}{\includegraphics[width=.075\linewidth,height=1cm]{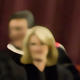}}&
\multicolumn{2}{c}{\includegraphics[width=.15\linewidth,height=1cm]{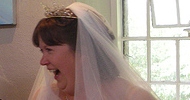}}&
\multicolumn{1}{c}{\includegraphics[width=.075\linewidth,height=1cm]{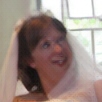}}&
\multicolumn{1}{c}{\includegraphics[width=.075\linewidth,height=1cm]{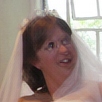}}&
\multicolumn{1}{c}{\includegraphics[width=.075\linewidth,height=1cm]{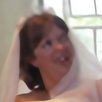}}
\\ \\
\multicolumn{2}{c}{\includegraphics[width=.15\linewidth,height=1cm]{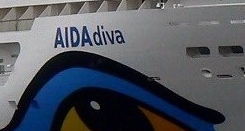}}&
\multicolumn{1}{c}{\includegraphics[width=.075\linewidth,height=1cm]{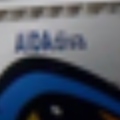}}&
\multicolumn{1}{c}{\includegraphics[width=.075\linewidth,height=1cm]{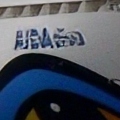}}&
\multicolumn{1}{c}{\includegraphics[width=.075\linewidth,height=1cm]{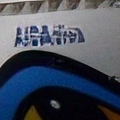}}&
\multicolumn{2}{c}{\includegraphics[width=.15\linewidth,height=1cm]{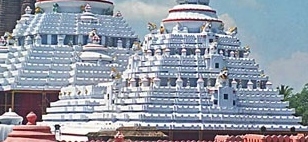}}&
\multicolumn{1}{c}{\includegraphics[width=.075\linewidth,height=1cm]{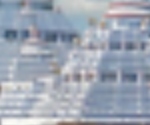}}&
\multicolumn{1}{c}{\includegraphics[width=.075\linewidth,height=1cm]{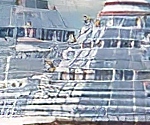}}&
\multicolumn{1}{c}{\includegraphics[width=.075\linewidth,height=1cm]{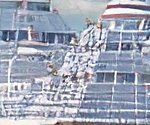}}
\\ 
\multicolumn{2}{c}{\includegraphics[width=.15\linewidth,height=1cm]{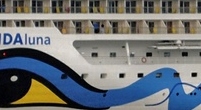}}&
\multicolumn{1}{c}{\includegraphics[width=.075\linewidth,height=1cm]{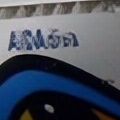}}&
\multicolumn{1}{c}{\includegraphics[width=.075\linewidth,height=1cm]{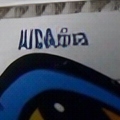}}&
\multicolumn{1}{c}{\includegraphics[width=.075\linewidth,height=1cm]{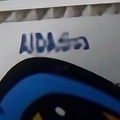}}&
\multicolumn{2}{c}{\includegraphics[width=.15\linewidth,height=1cm]{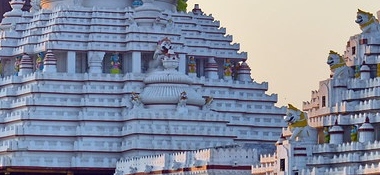}}&
\multicolumn{1}{c}{\includegraphics[width=.075\linewidth,height=1cm]{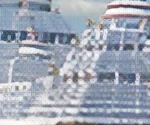}}&
\multicolumn{1}{c}{\includegraphics[width=.075\linewidth,height=1cm]{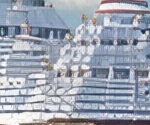}}&
\multicolumn{1}{c}{\includegraphics[width=.075\linewidth,height=1cm]{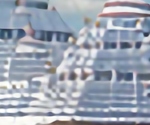}}
\\ \\
\multicolumn{2}{c}{\includegraphics[width=.15\linewidth,height=1cm]{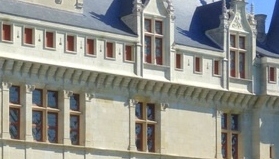}}&
\multicolumn{1}{c}{\includegraphics[width=.075\linewidth,height=1cm]{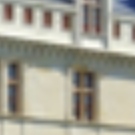}}&
\multicolumn{1}{c}{\includegraphics[width=.075\linewidth,height=1cm]{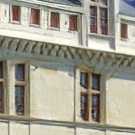}}&
\multicolumn{1}{c}{\includegraphics[width=.075\linewidth,height=1cm]{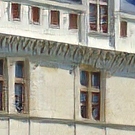}}&
\multicolumn{2}{c}{\includegraphics[width=.15\linewidth,height=1cm]{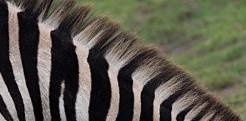}}&
\multicolumn{1}{c}{\includegraphics[width=.075\linewidth,height=1cm]{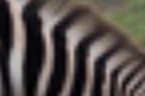}}&
\multicolumn{1}{c}{\includegraphics[width=.075\linewidth,height=1cm]{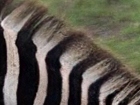}}&
\multicolumn{1}{c}{\includegraphics[width=.075\linewidth,height=1cm]{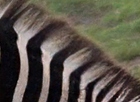}}
\\ 
\multicolumn{2}{c}{\includegraphics[width=.15\linewidth,height=1cm]{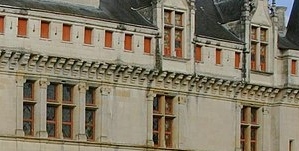}}&
\multicolumn{1}{c}{\includegraphics[width=.075\linewidth,height=1cm]{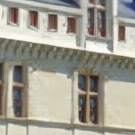}}&
\multicolumn{1}{c}{\includegraphics[width=.075\linewidth,height=1cm]{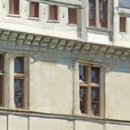}}&
\multicolumn{1}{c}{\includegraphics[width=.075\linewidth,height=1cm]{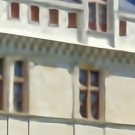}}&
\multicolumn{2}{c}{\includegraphics[width=.15\linewidth,height=1cm]{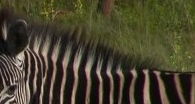}}&
\multicolumn{1}{c}{\includegraphics[width=.075\linewidth,height=1cm]{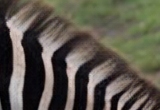}}&
\multicolumn{1}{c}{\includegraphics[width=.075\linewidth,height=1cm]{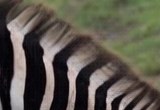}}&
\multicolumn{1}{c}{\includegraphics[width=.075\linewidth,height=1cm]{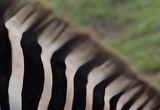}}
\\ 
\end{tabular}
\caption{Qualitative comparisons of our results with ESRGAN \cite{wang2018esrgan}, RankSRGAN \cite{zhang2019ranksrgan}, SRNTT \cite{zhang2019image}, and DATSR \cite{cao2022reference}.}
\label{qualitative_results}
\end{figure}

\textbf{Qualitative results.} Fig \ref{qualitative_results} shows qualitative comparisons with SOTA methods. We compare DARTS with ESRGAN \cite{wang2018esrgan}, RankSRGAN \cite{zhang2019ranksrgan}, SRNTT \cite{zhang2019image} and DATSR \cite{cao2022reference}. The outputs of DARTS show smooth images with fewer artifacts. As shown in the top left example, our model has reconstructed the face without mixing neighboring features. As shown in the second row of examples, DARTS can reconstruct the word "AIDA" more accurately than competing models. It can also reconstruct the fine features of the buildings without mismatching features and their relative positions. Overall, DARTS is able to preserve the integrity of global geometry while also performing fine-grained texture reconstruction.
\begin{table}[ht!]
\centering
\begin{tabular}{ c|cc } 
\hline
Methods & \#params & FLOPs \\
\hline
SRNTT-\textit{rec} \cite{zhang2019image}& 5.75M & 4.1G \\ 
TTSR-\textit{rec} \cite{yang2020learning}& 6.2M & 32.93G \\ 
$C2$-Matching-\textit{rec} \cite{jiang2021robust}& 8.9M & - \\ 
DATSR-\textit{rec} \cite{cao2022reference} & 18.0M  & - \\ 
DARTS-\textit{rec} & 22.29M & 24.8G \\ 
\hline
\end{tabular}
\caption{Comparison of the number of network parameters and FLOPs. \textit{`-rec'} denotes only the reconstruction ($l_1$-norm) loss. Compute information not available for $C^2$-matching and DATSR due to an inability to compile required libraries on the available hardware.}
\label{FLOPs}
\end{table}
\begin{table}[t]
\centering
\begin{tabular}{ c|ccccc} 
\hline
 & Self-attn & Cross-attn & Gating-attn & PSNR/SSIM\\
\hline
a) & \textcolor{green}{\cmark} & \textcolor{green}{\cmark} & \textcolor{green}{\cmark} & \textbf{26.4} / \textbf{.781} \\ 
b) & \textcolor{green}{\cmark} & \textcolor{green}{\cmark} & \textcolor{red}{\xmark} & 26.1 / .771 \\ 
c) & \textcolor{green}{\cmark} & \textcolor{red}{\xmark} & \textcolor{red}{\xmark} & 26.08 / .77\\ 
d) & \textcolor{red}{\xmark} & \textcolor{green}{\cmark} & \textcolor{red}{\xmark} & 25.96 / .768\\ 
\hline
\end{tabular}
\caption{Gating and two-stream attention work well together. We run ablation studies on DARTS-\textit{rec} trained for 50 epochs on the CUFED5 dataset. a) shows results using the full DARTS architecture, b) shows the performance when the gating parameter is frozen, c) uses only self-attention, and d) uses only cross-attention.}
\label{ablation_study}
\end{table}

\textbf{Ablation studies.} We perform three ablative studies as shown in Table~\ref{ablation_study}. These ablations show the individual effects of gating attention, self-attention and cross-attention. \textit{Gating attention} consists of both self-attention and cross-attention blocks that attend to the queries and keys from $I_{LR}$, as well as the queries from $I_{LR}$ and keys from $I_{ref}$ in each transformer block. The value vector is extracted from the $I_{ref}$ distribution. When gating attention is removed for ablation b), the $\lambda$ gating parameter is frozen at 0, paying equal attention to both self- and cross-attention scores. \textit{Self-attention} consists of a single self-attention block that only attends to the query, key and value feature vectors from the $I_{LR}$ distribution in each transformer block. This attention structure is very similar to the double attention introduced in \cite{zhang2022styleswin}. \textit{Cross-attention} consists of a single cross-attention block that only attends to the queries from $I_{LR}$, and keys and values from $I_{ref}$ in each transformer block. This attention structure is similar to vision and linguistic tasks proposed in \cite{lu2019vilbert}.

\begin{figure}[htb!]
\centering
\scriptsize
\begin{tabular}{cccc}
\multicolumn{1}{c}{\includegraphics[width=.25\linewidth,height=2.5cm]{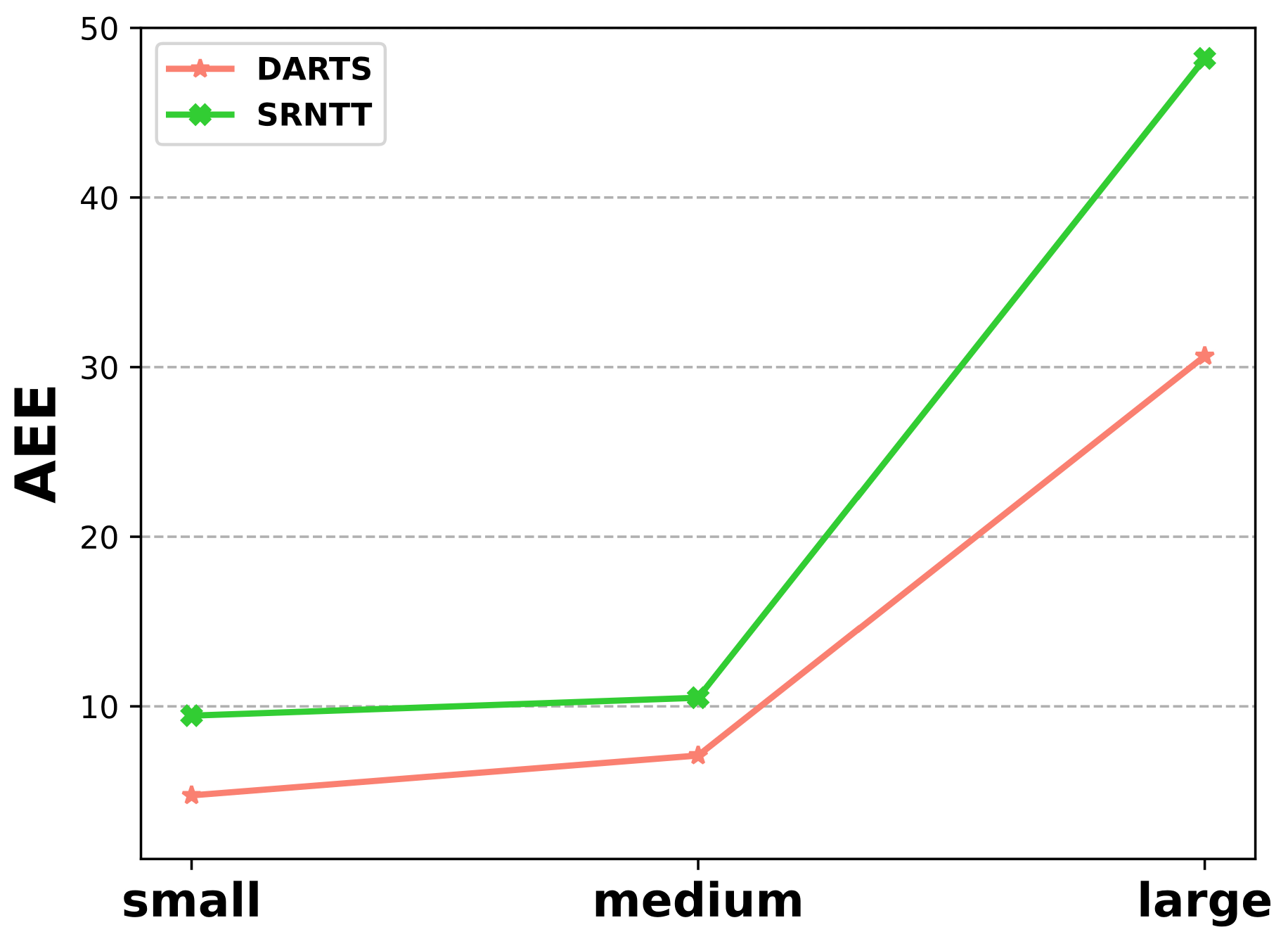}}&
\multicolumn{1}{c}{\includegraphics[width=.225\linewidth,height=2.5cm]{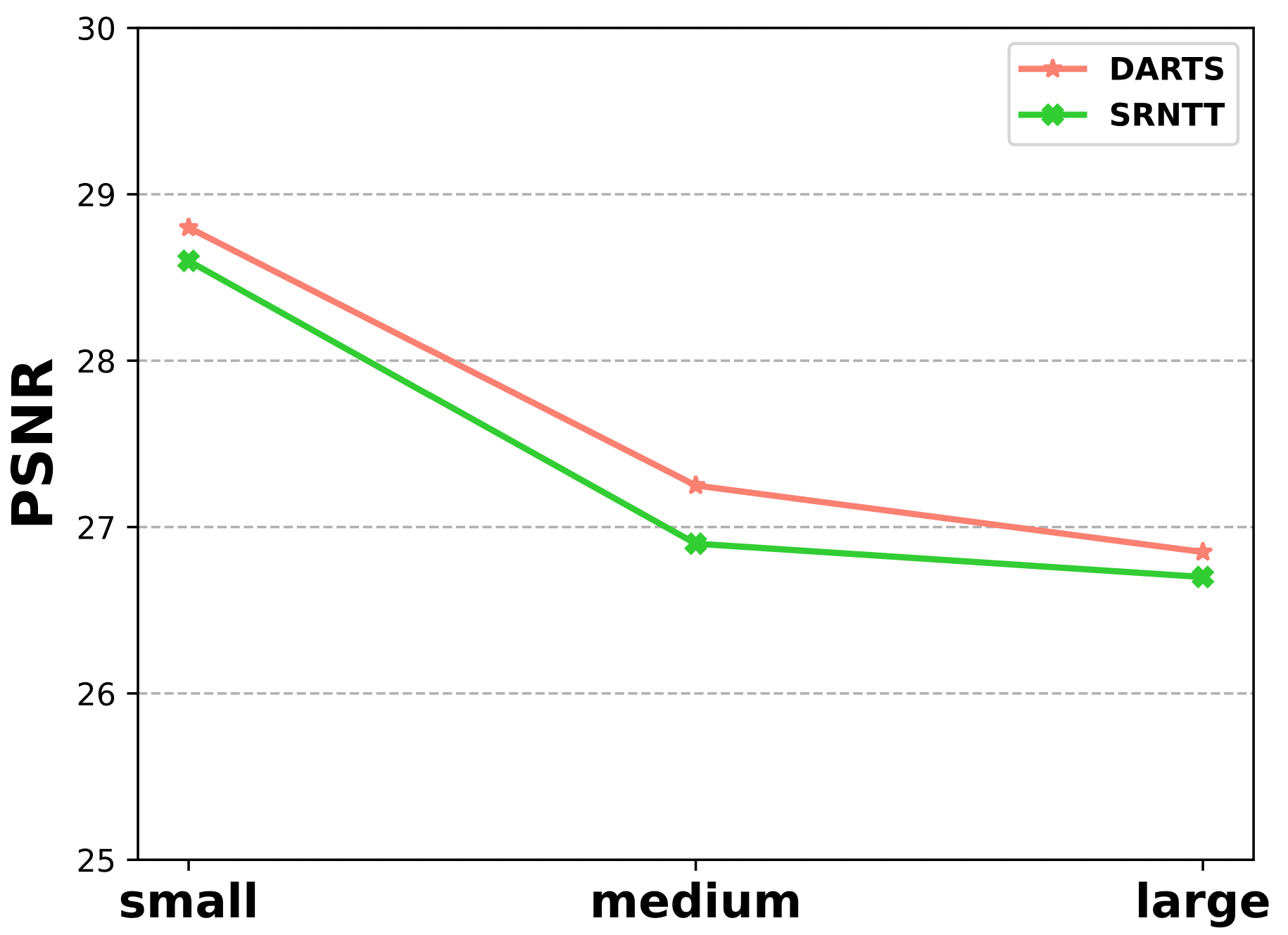}}&
\multicolumn{1}{c}{\includegraphics[width=.225\linewidth,height=2.5cm]{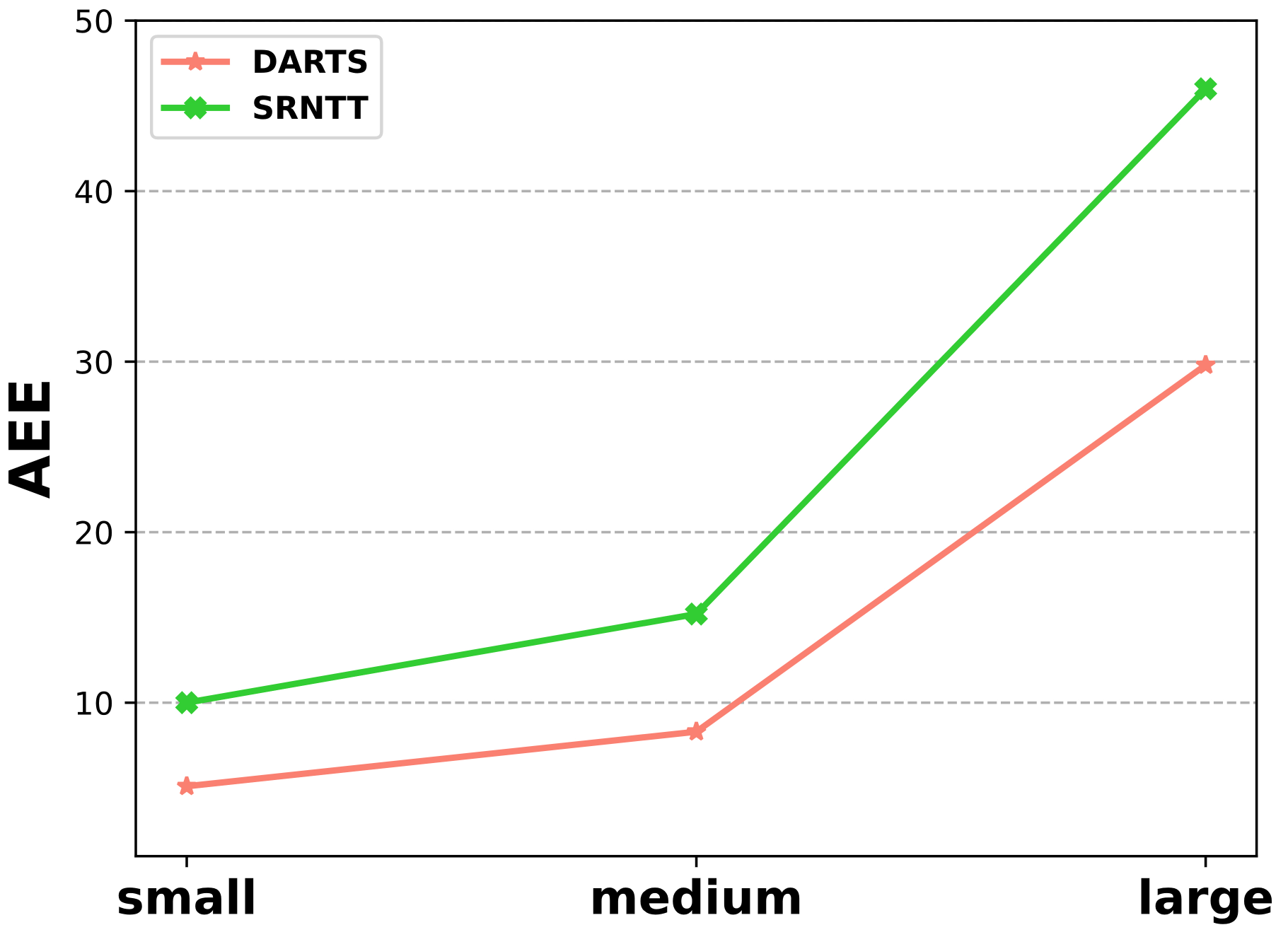}}&
\multicolumn{1}{c}{\includegraphics[width=.225\linewidth,height=2.5cm]{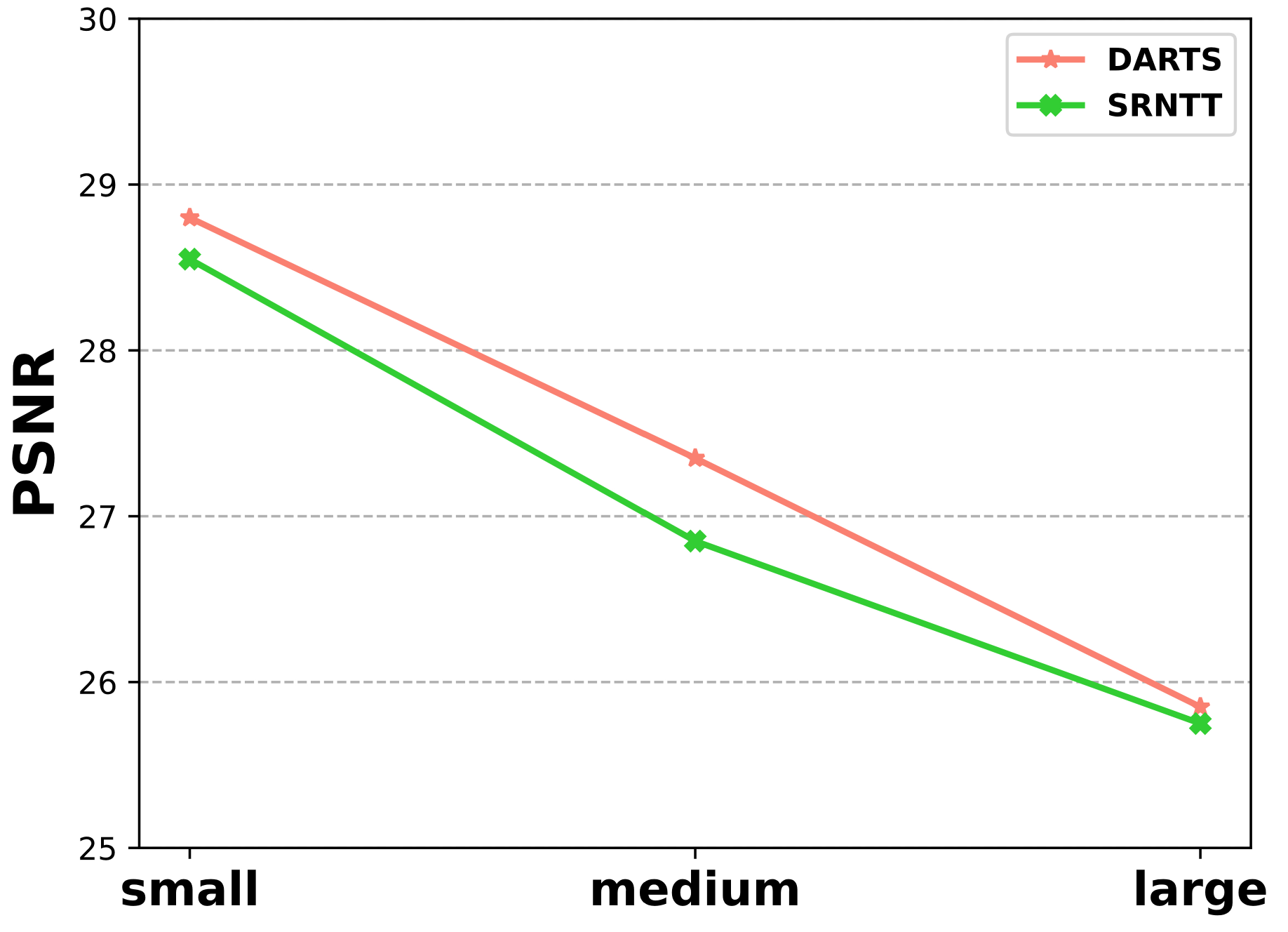}}\\
\multicolumn{2}{c}{a) Robustness to scale transformation} & \multicolumn{2}{c}{b) Robustness to rotation transformation}
\end{tabular}
\caption{DARTS shows more robustness than SRNTT \cite{zhang2019image} to scale (a) and rotation (b) across three levels of augmentation (small, medium, large), as measured by AEE (lower is better) and PSNR (higher is better).}
\label{graphs}
\end{figure}
\textbf{Scale and Rotation Invariance.} To demonstrate the robustness of our model to scale and rotation, we perform an analysis similar to the analyses conducted in \cite{jiang2021robust}. We rebuild the CUFED5 dataset with small, medium and large degrees of scale and rotation transformations. We use the scaled and rotated input images as reference images only at inference. We use Average End-to-point Error (AEE) and PSNR to assess the accuracy of matching correspondences and image restoration performance, respectively. Figure \ref{graphs} demonstrates the invariance of our model to the scale and rotation of objects during inference versus SRNTT \cite{zhang2019image}. As the degree of transformation increases, the AEE increases as well indicating more mismatched reconstructed features. Based on AEE, our model exhibits performance which is superior to SRNTT, even though we did not use large transformations during training. Based on PSNR, the restoration performance of DARTS is more robust than that of SRNTT as well.

\textbf{Limitations.} The main limitation of DARTS is the memory footprint of the model compared to previous architectures. Table~\ref{FLOPs} shows that DARTS has more parameters than SRNTT, TTSR, $C^2$-matching, and DATSR. However, the large memory requirement is largely due to the fact that attention is computed across four windows at once. This limits our experiments on consumer hardware to small batch sizes, as each GPU is only able to perform forward and backward passes on one sample at a time. It also limits the size of HR reference images, as mentioned in Section~\ref{sec:implementation}. This could potentially impact the performance of the model as it is sometimes necessary to crop the reference image into smaller patches.

\section{Conclusion}
In this work we proposed DARTS, a transformer model for reference-based image super-resolution. This proposed model consists of a two-stream architecture which simultaneously attends to the low-resolution input image and the high-resolution reference image, combining self-attention and cross-attention using a gating-attention strategy. The architecture is conceptually simple and consists of a single module which can be trained end-to-end in a single training stage. Quantitative and qualitative evaluations show that the proposed method is competitive with prior state-of-the-art methods which use a complex, multi-stage process with multiple submodules. It is also shown to be less sensitive to scale and rotation transformations of the reference images. DARTS represents a new direction for reference-based image super-resolution which is consolidated in the attention mechanism, as an alternative to more complicated SOTA methods based on $C^2$-matching.
\medskip
{\small
\bibliographystyle{abbrvnat}
\bibliography{neurips_2023.bib}
}




\end{document}